\documentclass[letterpaper]{article} 
\usepackage{aaai24}  
\usepackage{times}  
\usepackage{helvet}  
\usepackage{courier}  
\usepackage[hyphens]{url}  
\usepackage{graphicx} 
\usepackage{natbib}  
\usepackage{caption} 
\usepackage{algorithm}
\usepackage{algorithmic}
\usepackage{soul}
\usepackage{amsmath}
\usepackage{amsthm}
\usepackage{booktabs}
\usepackage{multirow}
\usepackage{color}
\usepackage{placeins}
\usepackage{amssymb}
\usepackage{subcaption}
\usepackage{algorithm}
\usepackage{algorithmic}
\usepackage{cleveref}
\usepackage{lipsum}
\usepackage{pifont}
\usepackage{utfsym}
\usepackage{bbding}

\usepackage{newfloat}
\usepackage{listings}
\DeclareCaptionStyle{ruled}{labelfont=normalfont,labelsep=colon,strut=off} 
\floatstyle{ruled}
\newfloat{listing}{tb}{lst}{}
\floatname{listing}{Listing}
\newcommand*\samethanks[1][\value{footnote}]{\footnotemark[#1]}
\title{SphereDiffusion: Spherical Geometry-Aware Distortion Resilient Diffusion Model}
\author {
    Tao Wu \textsuperscript{\rm 1}\equalcontrib ,
    Xuewei Li \textsuperscript{\rm 1}\equalcontrib ,
    Zhongang Qi \textsuperscript{\rm 2}\thanks{Corresponding author.},
    Di Hu\textsuperscript{\rm 3},
    Xintao Wang\textsuperscript{\rm 2},
    Ying Shan\textsuperscript{\rm 2},
    Xi Li \textsuperscript{\rm 1,\rm 4}\samethanks[2]
}
\affiliations {
    \textsuperscript{\rm 1}College of Computer Science and Technology, Zhejiang University \\
    \textsuperscript{\rm 2}ARC Lab, Tencent PCG \\
    \textsuperscript{\rm 3}Gaoling School of Artificial Intelligence, Renmin University of China\\
    \textsuperscript{\rm 4}Zhejiang – Singapore Innovation and AI Joint Research Lab, Hangzhou\\
    taowucs@zju.edu.cn, xueweili@zju.edu.cn, zhongangqi@tencent.com, dihu@ruc.edu.cn, xintaowang@tencent.com, yingsshan@tencent.com, xilizju@zju.edu.cn
}

\usepackage{bibentry}

\begin{document}
\maketitle

\begin{abstract}
Controllable spherical panoramic image generation holds substantial applicative potential across a variety of domains.
However, it remains a challenging task due to the inherent spherical distortion and geometry characteristics, resulting in low-quality content generation.
In this paper, we introduce a novel framework of SphereDiffusion to address these unique challenges, for better generating high-quality and precisely controllable spherical panoramic images.
For the spherical distortion characteristic, we embed the semantics of the distorted object with text encoding, then explicitly construct the relationship with text-object correspondence to better use the pre-trained knowledge of the planar images.
Meanwhile, we employ a deformable technique to mitigate the semantic deviation in latent space caused by spherical distortion.
For the spherical geometry characteristic, in virtue of spherical rotation invariance, we improve the data diversity and optimization objectives in the training process, enabling the model to better learn the spherical geometry characteristic.
Furthermore, we enhance the denoising process of the diffusion model, enabling it to effectively use the learned geometric characteristic to ensure the boundary continuity of the generated images.
With these specific techniques, experiments on Structured3D dataset show that SphereDiffusion significantly improves the quality of controllable spherical image generation and relatively reduces around 35\% FID on average.
\end{abstract}
\section{Introduction}
Spherical panoramic images, also known as $360^{\circ}$ panoramic images or omnidirectional panoramic images, are used in various domains such as autonomous driving~\cite{de2018eliminating,ma2021densepass,summaira2021recent}, virtual reality~\cite{xu2021spherical,ai2022deep}, etc. 
Numerous studies~\cite{yan2022horizon, hara2021spherical, akimoto2022diverse, li2011graph} have been proposed for the synthesis of spherical panoramic images, with a primary focus on reconstructing scenes from narrow field of view (NFOV) images. 
However, these generation methods often produce images of inferior quality and lack controllability, which are crucial in real applications. 

\begin{figure}[tb]
    \centering
    \includegraphics[width=1.0\linewidth]{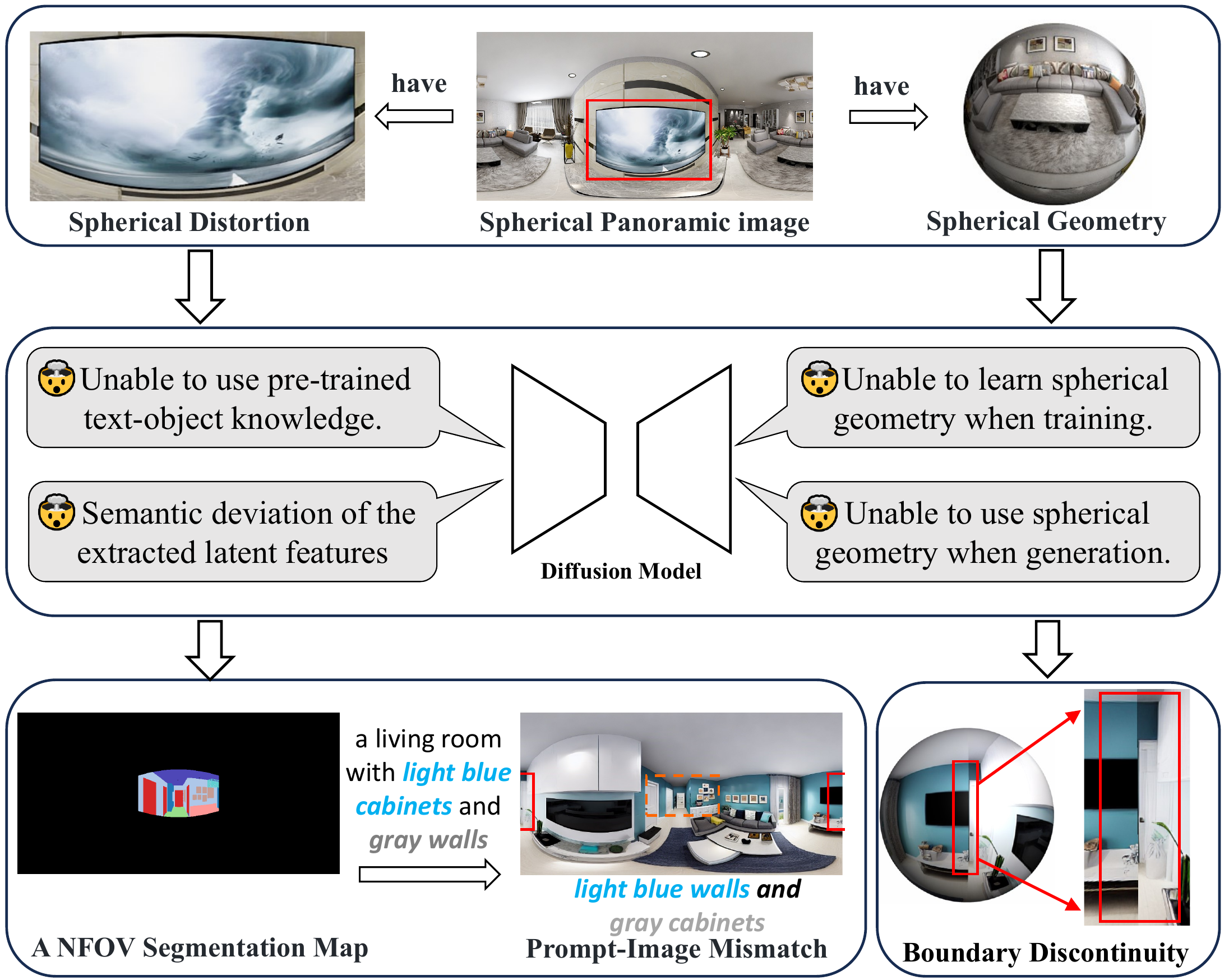}
    \caption{The characteristics of spherical panoramic images and the impact of these characteristics on existing controllable generation methods.}
    \label{fig:first}
\end{figure}
In contrast, extensive works~\cite{zhang2023adding,mou2023t2i} have greatly succeeded in controllable high-quality planar image generation.
Most of the existing works resort to fine-tuning the pre-trained large-scale diffusion models to adapt to different application scenarios. 
However, such a paradigm falls short of expectations for generating the spherical panoramic images, since simply fine-tuning cannot capture the unique characteristics of the spherical panoramic images.  

Two characteristics of spherical panoramic images are essentially different from the planar images: \textbf{spherical distortion} and \textbf{spherical geometry}. 
As shown in ~\Cref{fig:first}, on the one hand, spherical distortion mainly refers to the deformation of objects.
Existing controllable generation models are primarily designed and pre-trained based on planar images. 
Thus, the text-object correspondence knowledge stored in these pre-trained weights cannot be effectively utilized for spherical panoramic images due to the significant deformation of distorted objects.
At the same time, spherical distortion makes it difficult to extract effective features of spherical panoramic images, resulting in semantic deviation when depicting image content.
Accordingly, text prompts often fail to correctly guide the generation process, leading to the mismatch between text guidance and generated visual content.
On the other hand, spherical geometry means that the visual content of a spherical panoramic image is a projection of the 3D world on a sphere.
Such sphere structure shares 3D geometric attributes such as spherical rotation invariance and non-boundary property.
Current controllable planar generative models lack geometry-aware training design, resulting in the difficulty of effectively incorporating the spherical geometry characteristic.
As a consequence, improving the quality of the generated content using spherical geometry becomes a challenge.
Practically, preventing the model from generating spherical images from a global perspective leads to issues like boundary discontinuity.
Considering the above issues, one question naturally arises: How can we enable the model to learn and utilize the characteristics of spherical images, then enhance the quality of controllable spherical panoramic image generation?

In this work, we propose the SphereDiffusion framework, which targets to generate high-quality and precisely controllable spherical panoramic images from single NFOV segmentation maps and text prompts. 
To solve the above issues, we impose the two characteristics of spherical panoramic images into the model design, as well as the training and inference process.
Concretely, for spherical distortion characteristic, we introduce Distortion-Resilient Semantic Encoding (DRSE) to enhance the utilization of pre-trained knowledge.
It embeds the text semantics into distorted objects, aligning text-object correspondence knowledge of pre-trained planar image generation models and the objects in the spherical panoramic image.
Meanwhile, we also introduce a Deformable Distortion-aware Block (DDaB) constructed based on deformable convolution to relieve semantic deviation. 
The deformability of DDaB helps the model extract effective features from distorted objects with different deviations in spherical panoramic images. 

For spherical geometry characteristic, we aim to let the model adequately learn and use it, then improve both the training process and generation process.
On the one hand, we propose Spherical Geometry-aware (SGA) Training, enabling the model to better learn the spherical geometry characteristic. 
It contains two modules: Spherical Reprojection and Spherical SimSiam Contrastive Learning.
Spherical Reprojection applies spherical rotation invariance to the training data, enabling the model to better learn spherical geometry through data diversity.
Spherical SimSiam Contrastive Learning ensures spherical rotation invariance in the latent space, increasing the spherical robustness of models at the optimization objective.
On the other hand, we introduce SGA Generation, which allows the model to better use the spherical geometry characteristic to improve the generation process.
By incorporating spherical rotation invariance into the generation process, we iteratively rotate the intermediate results from the previous denoising step to connect the content located at the two ends of the intermediate results.
In this way, the boundary connectivity of the generated image is improved.
Our contributions are summarized as follows:
\begin{itemize}
    \item We proposea novel framework for controllable spherical panoramic image generation, which takes both spherical geometry and image distortion into consideration.
    \item We propose DRSE and DDaB to deal with spherical distortion, enabling the model to better use the pre-trained knowledge and reduce the semantic deviation in latent space caused by spherical distortion.
    \item We introduce SGA Training to make models learn spherical geometry from both data diversity and optimization objectives. We also propose SGA Generation to improve the denoising process of the diffusion model.
\end{itemize}
Experimental results on the Structured3D dataset~\cite{Structured3D} demonstrate that our method can significantly improve the quality of controllable spherical image generation and relatively reduces around 35\% FID on average compared to previous methods.
\section{Related Work}
\subsection{Conditional Diffusion Probabilistic Model}

Diffusion model~\cite{sohl2015deep,dhariwal2021diffusion} is a generative probability model, which has attracted many researchers' attention because of its high-quality generative results.
Diffusion models can successfully perform conditional image generation when trained with guidance such as semantic layout or class labels~\cite{zheng2022entropy,ramesh2021dalle1,saharia2022image,ho2022classifier,zheng2023layoutdiffusion,xue2023freestyle,chen2014ranking}.
A notable example of conditional diffusion models is recent text-to-image diffusion models, which have showcased unprecedented synthetic capabilities ~\cite{nichol2021glide,saharia2022photorealistic,sheynin2022knn,jiang2019learning}. 
Recently, many methods have been observed to enhance user controllability.
Existing methodologies can be broadly bifurcated into two primary strategies:
(i) Approaches that integrate explicit control by incorporating additional guiding signals into the model ~\cite{avrahami2022spatext,rombach2022high,brooks2023instructpix2pix}.
However, these studies require costly training on meticulously curated datasets. 
(ii) Many methods have been proposed to implicitly control the content generated by manipulating the generation process of a pre-trained model ~\cite{mokady2023null,kong2023leveraging}
or by conducting lightweight model fine-tuning ~\cite{ruiz2023dreambooth,kawar2023imagic,zhang2023adding}. 
Most of these methods only require minimal training overhead, making them the mainstream approach for controllable generation.

\subsection{Spherical Panoramic Image Generation}
Current spherical panoramic image generation techniques can be divided into two categories: GAN-based generative models and diffusion-based generative models.
Kimura et al.~\cite{kimura2018extvision} presented a peripheral image generation technique based on pix2pix ~\cite{isola2017image}. 
However, the FOV used to generate image was constrained. 
Sumantri et al.~\cite{sumantri2020360} advanced a spherical image generation technique based on pix2pixHD ~\cite{wang2018high}, which required a collection of images taken from various directions as input. 
Hara et al.~\cite{hara2021spherical} present a novel method to generate spherical images from a single NFOV image by controlling the degree of freedom of generated regions using scene symmetry.
Along with the development of the diffusion model, some panoramic image generation methods based on the diffusion model have emerged.
Bar-Tal et al.~\cite{bar2023multidiffusion} define a new generation process to generate panoramas, which is composed of several reference diffusion generation processes bound together with a set of shared parameters or constraints and without further training or fine-tuning. 
Zhang et al.~\cite{zhang2023diffcollage} propose a combinatorial diffusion model that can take advantage of a trained diffusion model based on factor graph representations to generate spherical panoramic images.
Li et al.~\cite{li2023panogen} use recursive overpainting on generated images to create consistent spherical panoramic views by conditioning text descriptions.
\section{Preliminaries}
\label{sec:preliminaries}
\subsection{Latent Diffusion Models}

Diffusion models are probabilistic models designed to learn a data distribution $p(x)$ by gradually denoising a normally distributed variable and can be interpreted as a sequence of denoising autoencoders $\epsilon_\theta\left(x_t, t\right)$. 
They are trained to predict denoised versions of their inputs $x_t$, where $x_t$ is a noisy variant of the input $x$. 
Latent diffusion models (LDMs)~\cite{rombach2022high} employ a two-stage approach to train diffusion models directly in high-resolution pixel space with acceptable computational cost.  
First, a learnable autoencoder (consisting of an encoder $\textbf{E}$ and a decoder $\textbf{D}$) is trained to compress the image into a smaller latent space representation.
Then, a diffusion model of representations $z = \textbf{E}(x)$ is trained instead of a diffusion model of images $x$.
Moreover, in the forward process, LDM incrementally adds noise to $z$ to get $z_t$ and performs denoising to predict $z$ in the reverse process.
New images can be generated by sampling a representation $\tilde{z}$ from the diffusion model and subsequently decoding it into an image using the learned decoder $\tilde{x} = \textbf{D}(\tilde{z})$. 
During training, the loss is defined as follows:
\begin{equation}
L_{LDM}=\mathbb{E}_{z_0, \epsilon \sim \mathcal{N}(0,1), t}\left[\left|\epsilon-\epsilon_{\theta}\left(z_t, t\right)\right|_2^2\right]. 
\end{equation}

\subsection{Controllabel Image Synthesis Diffusion Models}
Controllable image synthesis diffusion models allow the creation of diverse images based on text instructions or guidance from a reference image. 
ControlNet, a trainable adaptor, is specifically designed to function in tandem with Stable Diffusion, which is a representative work of this field. 
A simple network $\mathcal{F}_{hint}$ is first used to downsample the input control image $c$ to the same size as the input vector $z$ in the latent space of Stable Diffusion, yielding $C_{latent}$. 
Subsequently, Controlnet uses its control branch $\mathcal{F}_{c}$ to process $C_{latent}$, resulting in multi-scale features $F_{c}=[F_{c}^{1},F_{c}^{2},...,F_{c}^{n}]$.
These features are then added to the features of the same resolution at the corresponding positions in the middle block and the decoder block of the U-Net structure in Stable Diffusion. 
This process effectively controls the generation of Stable Diffusion. 
During training, the main training constraint is defined as follows:
\begin{equation}
L_{C}=\mathbb{E}_{\boldsymbol{z}_0, t, \boldsymbol{c}_t, c, \epsilon \sim \mathcal{N}(0,1)}\left[\left| \epsilon-\epsilon_\theta\left(z_t, t, \boldsymbol{c}_t, c\right)\right|_2^2\right].
\end{equation}
In this paper, we adopt ControlNet as our baseline. 

\section{Method}
\label{sec：method}
In this section, we present the fundamental idea and detailed design of SphereDiffusion. 
First, we provide an overview of controllable spherical panoramic image generation. 
Second, we describe our solution to spherical distortion. 
Finally, we introduce our strategy to allow the model to learn and utilize the characteristic of spherical geometry better.
 
\subsection{Overview}
\label{ssec:overview}
SphereDiffusion generates high-quality controllable spherical panoramic images $x$ which simultaneously conform to a corresponding text prompt $C_{text}$ and an NFOV segmentation map $C_{mask}$.
The foundational ControlNet serves as the baseline for this process.
In order to improve the quality of controllable spherical panoramic image generation, SphereDiffusion needs to deal with two main characteristics, spherical distortion and spherical geometry. 

Spherical distortion causes a certain category of objects in different positions in the spherical panoramic image to show significant and different shape changes compared with the planar image.
This poses challenges to the model to effectively utilize the text-object correspondence knowledge stored in pre-trained weights and extract effective features of distorted objects. 
First, to better use the pre-trained knowledge of planar images, we propose our Distortion-Resilient Semantic Encoding (DRSE), to align the input condition to the pre-trained text-object correspondence knowledge.
In addition, to deal with the different shape changes of objects at different locations of a spherical panoramic image, we propose our Deformable Distortion-aware Block (DDaB). 

Spherical geometry has several unique properties, such as spherical rotation invariance and non-boundary property. 
To enable the model to learn and use spherical geometry, we introduce SGA Training and SGA Generation during the training and generation processes, respectively.
SGA Training enhances data diversity and optimization objectives during the training process by employing Spherical Reprojection and Spherical SimSiam Contrastive Learning, respectively. 
This approach enables the model to learn the spherical geometry characteristic better.
Furthermore, SGA Generation uses learned geometric characteristics to enhance the boundary connectivity of spherical panoramic images to make the generated content continuous.

\begin{figure*}[tb]
    \centering
    \includegraphics[width=1.0\linewidth]{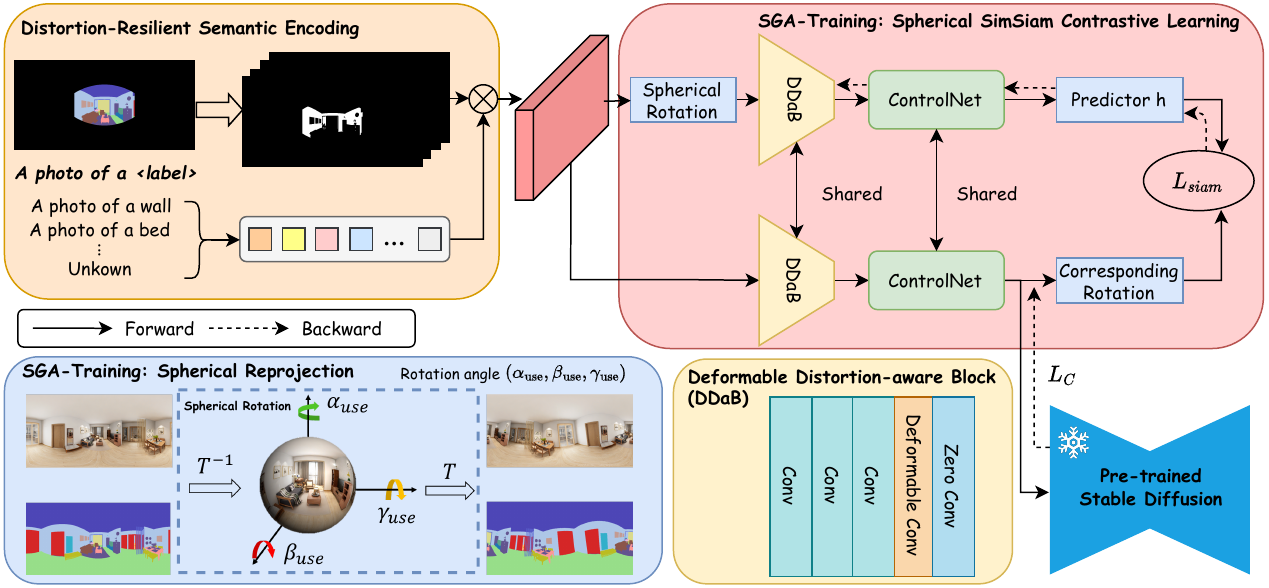}
    \caption{Overall review of SphereDiffusion. 
    (Upper left) Distortion-Resilient Semantic Encoding introduces category information into the representation of segmentation maps to alleviate the issue of text-image mismatch. 
    (Upper right) Spherical SimSiam Contrastive Learning is a part of SGA Training, which constructs contrastive learning in the latent space, equipping SphereDiffusion with spherical geometry at the objective function level. 
    (Lower left) Spherical Reprojection is a part of SGA Training at the data level, and Spherical Rotation serves as the foundation for SGA Training. 
    (Lower middle) DDaB with deformable convolution enhances the model's perceptual ability of spherical distortion.}
    \label{fig:pipline}
    \centering
\end{figure*}

\subsection{Spherical Distortion Properties Solution}
\label{ssec:sdpsolution}
As mentioned above, the impact of spherical distortion has two main aspects. 
First, to utilize the text-object correspondence knowledge stored in pre-trained weights, we replace the original RGB segmentation map with a segmentation map rich in semantic information.
Moreover, to reduce the semantic deviation in latent space, models should be specially designed to extract effective features of different locations of a spherical panoramic image differently and adaptively.
Inspired by Trans4PASS~\cite{zhang2022bending}, we improve the $F_{hint}$ by deformable technique through our Deformable Distortion-aware Block. 
\subsubsection{Distortion-Resilient Semantic Encoding}
\label{sssec:clip}
Distortion-Resilient Semantic Encoding starts from the perspective of input data, upgrading the connection between color information in the segmentation map and the generated objects to the connection between class semantic information in the segmentation map and the generated objects. 
This allows the model to better utilize the text-object correspondence knowledge stored in pre-trained weights.
Specifically, as shown in the upper left of ~\Cref{fig:pipline}, given one NFOV segmentation map $C_{mask}$, we first set the segmentation maps for the remaining positions in a newly introduced category, "Unknown". 
This results in our final input segmentation map, $C_{mask}^{\prime}$. 
Then, we downsample the segmentation map to the same resolution as the input vector $z \in \mathbb{R}^{C \times H \times W}$ in the latent space of Stable Diffusion.
According to the categories of the labels, we divide the entire image into $K$ two-dimensional binary masks $M = \left\{m_i \mid m_i \in[0,1]^{H \times W}\right\}_{i=1}^K$, where $K$ represents the total number of categories, including the newly added 'Unknown' category.
Subsequently, we construct the label texts using the prompt template `a photo of a \{label\}' for all categories $\mathcal{L} = \{ l_1,l_2,...,l_K \}$. These label texts are then encoded using the text encoder of CLIP~\cite{radford2021learning}, resulting in label embeddings $\mathcal{E}_{\text {label}} \in \mathbb{R}^{C_{\mathcal{E}} \times K}$.
Finally, we multiply the binary masks $M$ with the label embeddings $\mathcal{E}_{\text {label}}$, resulting in a per-pixel embedding $\mathcal{E}_{\text {pixel}} \in \mathbb{R}^{C_{\mathcal{E}} \times H \times W}$. 
We use $\mathcal{E}_{\text {pixel }}$ as the guiding input for the final model ($F_{CLIP}$ is the text encoder of the CLIP model, $\otimes$ is the matrix cross product):

\begin{gather}
    C_{mask}^{\prime} \rightarrow M = \left\{m_i \mid m_i \in[0,1]^{H \times W}\right\}_{i=1}^K, \\
    \mathcal{E}_{\text {label}} = F_{CLIP}(\mathcal{L}), \\
    \mathcal{E}_{\text {pixel}} = \mathcal{E}_{\text {label}} \otimes M.
\end{gather}
\begin{figure*}[tb]
    \centering
    \includegraphics[width=0.98\linewidth]{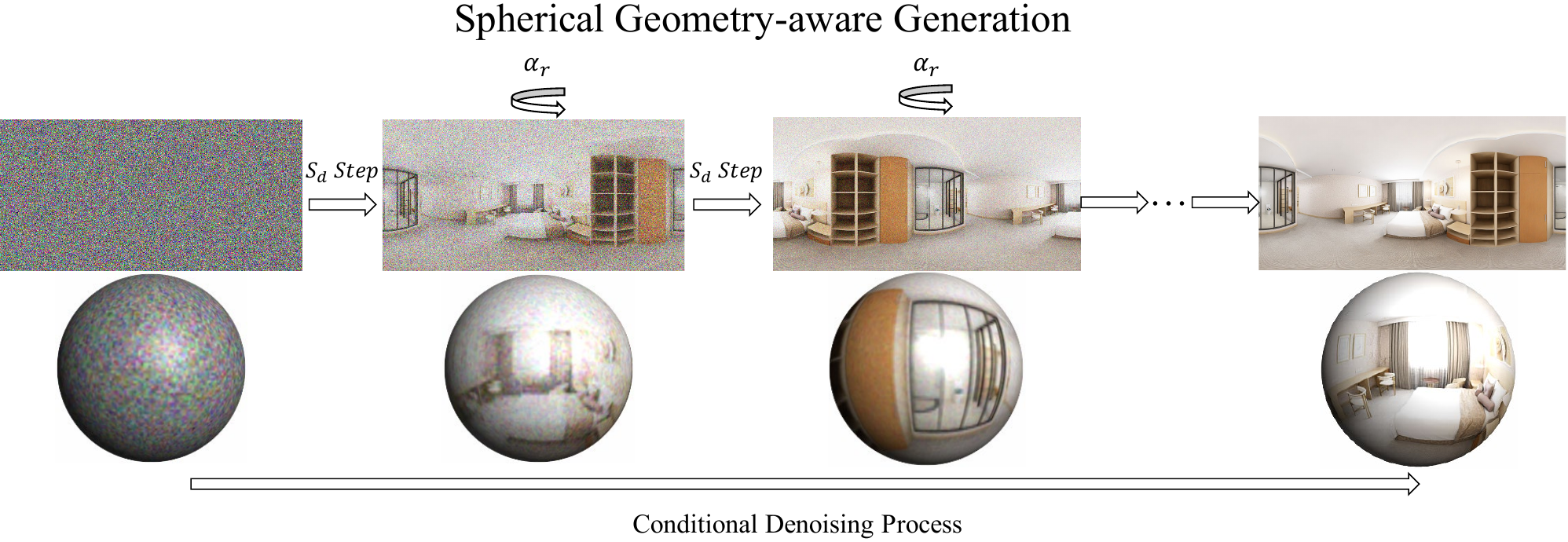}
    \caption{The processing of Spherical Geometry-aware Generation. During the generation process, we uniformly select K steps to rotate an angle $\alpha^{r}$ to enhance the boundary connectivity of the generated image.}
    \label{fig:sgag}
    \centering
\end{figure*}
\subsubsection{Deformable Distortion-aware Block}
\label{sssec:DDaB}
The Deformable Distortion-aware Block starts from the perspective of the model structure, introducing a deformable convolution into the model. 
This allows the model to better adapt and extract effective features from spherical images.
After obtaining $\mathcal{E}_{\text {pixel }}$, to align with the original design of ControlNet, we use a learnable block to transform $\mathcal{E}_{\text {pixel}}$ into an embedding with the same dimensions as $z$. 
To deal with the different shape changes of objects at different locations of a spherical panoramic image, we introduce deformable convolution within this block.
In detail, for each image, the offsets $\Delta_{(i,j)}$ of the $i^{th}$ row $j^{th}$ column pixel are defined as: 
\begin{equation}
  \label{eq:DPE}
  \Delta_{(i,j)}={\begin{bmatrix}
  \mathrm{min}(\mathrm{max}(\text{-}k_{D} \cdot H, g(f)_{(i,j)}),k_{D} \cdot H)\\
  \mathrm{min}(\mathrm{max}(\text{-}k_{D} \cdot W, g(f)_{(i,j)}),k_{D} \cdot W)
  \end{bmatrix}}, 
\end{equation}
where $g(\cdot)$ is the offset prediction function. 
The hyperparameter $k_{D}$ puts an upper bound on the learnable offsets $\Delta$.  
For implementation, as shown in the lower middle of ~\Cref{fig:pipline}, in block $\mathcal{F}_{hint}$, which consists of four convolutional layers and one zero convolution layer, we replace the fourth convolutional layer with a deformable convolutional layer.

\subsection{Spherical Geometry-aware Diffusion Model}
\label{ssec:sgaddpm}
We made improvements to both the training and inference process.
SGA Training fuses spherical geometry in data diversity and optimization objective. 
We adopt random rotations in 3D space to enhance data diversity. 
In terms of optimization objective, we propose Spherical SimSiam Contrastive Learning to make the extracted features equipped with spherical rotation invariance. 
we introduce SGA Generation, which allows the model to better use the spherical geometry characteristic to improve the generation process.
\subsubsection{Spherical Geometry-aware (SGA) Training}
\label{sssec:sgatrain}
Traditional training strategies treat the input as a planar image. 
It results in the model overfitting to images of a single projection way, thereby inhibiting the model's ability to learn the spherical geometry characteristic. 
Therefore, we introduce SGA training with the aim of enhancing the ability of the model's control branch to learn the spherical rotation invariance of spherical images. 
First, we introduce the spherical rotation.
As shown in the lower left of~\Cref{fig:pipline}, let $T$ denote the forward transformation of the Equirectangular Projection (ERP), which entails the conversion of spherical coordinates to planar coordinates. 
$T^{-1}$ signify the inverse one. 
Given an input panoramic image processed through ERP, we initially convert the image $I$ to spherical coordinates by applying the inverse ERP transformation. 
Subsequently, benefiting from ~\cite{li2023sgat4pass}, we employ a three-dimensional rotation matrix within the spherical coordinate system to execute a three-dimensional rotation. 
For a generic rotation in three-dimensional space, the angles of yaw, pitch, and roll are represented by $\alpha_{\mathrm{use}}$, $\beta_{\mathrm{use}}$, and $\gamma_{\mathrm{use}}$, respectively. 
The associated rotation matrix is denoted by $R(\alpha_{\mathrm{use}}, \beta_{\mathrm{use}}, \gamma_{\mathrm{use}})$. 
By multiplying $R$ with the data in the spherical coordinate system, we acquire the rotated data within the same coordinate system. 
Ultimately, we apply the ERP forward transformation to convert the rotated spherical coordinate system image into a panoramic image, thereby obtaining a specific rotated image of the real input of the model. 
The corresponding point in the input image of a pixel in the rotated image may not possess integer coordinates; thus, we choose the nearest pixel as its corresponding pixel. 
In summary, the rotation process of a spherical image $I$ can be defined as follows: $O_{3D}(I, \alpha_{\mathrm{use}}, \beta_{\mathrm{use}}, \gamma_{\mathrm{use}}) = T(R(\alpha_{\mathrm{use}}, \beta_{\mathrm{use}}, \gamma_{\mathrm{use}}) \cdot T^{-1}(I))$.
During training, two methods help the model learn the geometric property, including spherical rotation invariance.

Spherical Reprojection: Given a data pair $(x$, $C_{mask}$, $C_{text})$, we can rotate both $x$ and $C_{mask}^{\prime}$ by a random rotation angle chosen randomly within the maximum rotation angle $(\alpha_{d}, \beta_{d}, \gamma_{d})$ to obtain $x^{r}$ and $c_{mask}^{r}$, thereby generating more data $(x^{r}, C_{mask}^{r}, C_{text})$.
This method allows the model to learn geometric properties directly.

Spherical SimSiam Contrastive Learning: Benefiting from SimSiam~\cite{chen2021exploring}, we use $C_{latent}$ representing the input of the control branch $\mathcal{F}_{c}$ in ControlNet. 
As shown in the upper right part of~\Cref{fig:pipline}, we randomly rotate $C_{latent}$ using a random rotation $(\alpha_{use}, \beta_{use}, \gamma_{use})$ chosen randomly within the maximum rotation angle $(\alpha_{c}, \beta_{c}, \gamma_{c})$ to obtain a new view $C_{latent}^{r}=O_{3D}(C_{latent},\alpha_{use}, \beta_{use}, \gamma_{use})$. 
The encoders $\mathcal{F}_{c}$ of the two branches share the same weights. 
A prediction MLP head $h$ transforms the output of one view and matches it to the other view. 
We maximize the cosine similarity between the two branches as follows:
\begin{equation}
    \mathcal{D}\left(p_1, z_2\right)=-\frac{p_1}{\left\|p_1\right\|_2} \cdot \frac{z_2}{\left\|z_2\right\|_2},
\end{equation}
where $\|\cdot\|_2$ is $\ell_2$-norm, $p_1 = h(\mathcal{F}_{c}(C_{latent}))$, $z_2 = O_{3D}(\mathcal{F}_{c}(C_{latent}^{r}),\alpha_{use}, \beta_{use}, \gamma_{use})$.
Then we define a symmetrized loss as follows:
\begin{equation}
    L_{siam}=\frac{1}{2} \mathcal{D}\left(p_1, \operatorname{stop}\left(z_2\right)\right)+\frac{1}{2} \mathcal{D}\left(p_2, \operatorname{stop}\left(z_1\right)\right)
\end{equation}
where $\operatorname{stop}$ represents the stop-gradient operation and prevents a degenerate solution due to model collapse. 
We set our total loss as ($\lambda$ is a hyperparameter):
\begin{equation}
        L_{all} = L_{c}+ \lambda \cdot L_{siam}. 
\end{equation}
\subsubsection{Spherical Geometry-aware (SGA) Generation}
\label{sssec:sgatest}
During the generation process with diffusion models, the output is not produced in a single step; rather, it involves multiple iterations.
Therefore, inspired by the non-boundary property of the spherical panoramic image, we have also improved the generation process.
As shown in \Cref{fig:sgag}, assuming that we need $N$ steps $\{t^{1},t^{2}...,t^{N}\}$ to complete the generation of the diffusion model, we will uniformly select $K$ steps throughout the process $S = \{s^{1},s^{2}...,s^{K}\}$, $S_{d} = \frac{N}{K+1}$ steps between each step. 
When the current iteration step is $t \in S$, we simultaneously rotate the latent space vector and the guided segmentation map at an angle of $\alpha_{r} = \frac{360^{\circ}}{K}$.
Such an approach can enhance the boundary connectivity of the spherical image during the generation process.

\section{Experiments}
\subsection{Datasets, Protocols, and Evaluation Metrics}
We evaluated our model on the Structured3D dataset~\cite{Structured3D}, which provides 196k spherical panoramic images of 21,835 rooms in 3,500 scenes.
We use scene\_00000 to scene\_03249 for training, and scene\_03250 to scene\_03499 for testing.
Our experiments are conducted with a server with eight NVIDIA A100 GPUs, and training epochs are 20. 
The base model is Stable Diffusion 1.5, and text prompts are annotated with BLIP~\cite{li2022blip}. 
Following the settings in \cite{hara2021spherical}, during the training process, we extract an NFOV image from a spherical image with a field of view ranging from $30^{\circ}$ to $120^{\circ}$ and an aspect ratio of $2:1$.
Subsequently, the viewpoint direction was arbitrarily established on the sphere and projected onto the equirectangular image.
We set $(\alpha_{c}, \beta_{c}, \gamma_{c}) = (360^{\circ},3^{\circ},3^{\circ})$ and $(\alpha_{d}, \beta_{d}, \gamma_{d}) = (360^{\circ},10^{\circ},10^{\circ})$. 
$\lambda$ / $N$ / $K$ are set to 0.1 / 50 / 4, respectively.
We choose widely used metrics to evaluate image generation quality, including Fréchet Inception Distance (FID)~\cite{heusel2017gans}, spatial Fréchet Inception Distance (sFID)~\cite{nash2021generating}, and Inception Score (IS)~\cite{salimans2016improved}. 

\subsection{Performance Comparison}
\label{sssec:sgatest}
\begin{figure*}[tb]
    \centering
    \includegraphics[width=0.98\linewidth]{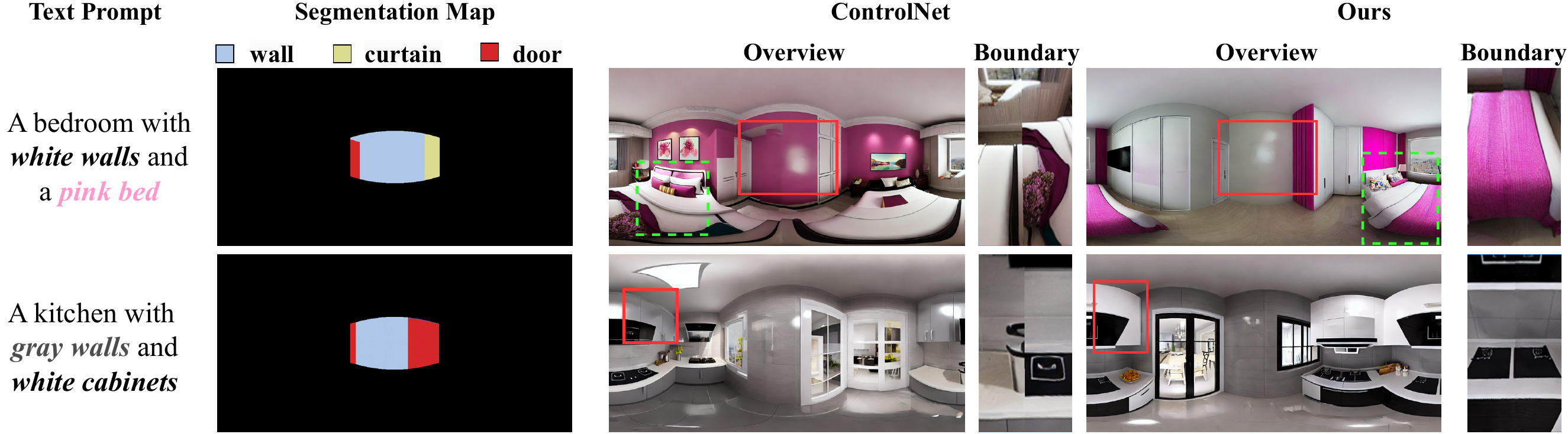}
    \caption{Visualization comparison of comparing SphereDiffusion with ControlNet. 
    The images generated by our SphereDiffusion are more closely aligned with the guidance provided by the segmentation maps and text prompts (highlighted by red line boxes and green dotted boxes). `Overview' is generated image, and `Boundary' displays the boundary of the generated image.}
    \label{fig:cmpsota}
    \centering
\end{figure*}
\begin{table}[]
    \centering
    \begin{tabular}{lcccc}
        \toprule
        Method     & FOV                  & FID$\downarrow$    & sFID$\downarrow$    & IS$\uparrow$    \\ \midrule
        ControlNet & \multirow{2}{*}{$30^{\circ}$}  & 44.801 & 174.841 & 3.006 \\
        Ours       &                      & 29.156 & 121.607 & 3.323 \\
        \midrule
        ControlNet & \multirow{2}{*}{$60^{\circ}$}  & 41.917 & 158.873 & 2.957 \\
        Ours       &                      & 26.262 & 111.318 & 3.325 \\
        \midrule
        ControlNet & \multirow{2}{*}{$90^{\circ}$}  & 39.450  & 142.747 & 2.954 \\
        Ours       &                      & 25.042 & 105.165 & 3.234 \\
        \midrule
        ControlNet & \multirow{2}{*}{$120^{\circ}$} & 35.690  & 123.075 & 2.853 \\
        Ours       &                      & 24.147 & 92.039  & 3.246 \\ \bottomrule
    \end{tabular}
    \caption{Comparison with the existing methods on Strcuture3D dataset. We use the same hyperparameter settings and number of training epochs for a fair comparison.}
    \label{tab:sota_comp}
\end{table}
\begin{table}[]
    \footnotesize
    \centering
    \begin{tabular}{cccccc}
    \toprule
    DRSE                 & DDaB         & SR                & SSCL                 & SGAG           & FID$\downarrow$      \\ \midrule
        \ding{55}                &   \ding{55}  &   \ding{55}       &     \ding{55}        &    \ding{55}     & 39.450               \\
    \ding{51}      &   \ding{55}  &   \ding{55}       &     \ding{55}        &         \ding{55}    & 38.805               \\
    \ding{51}                 & \ding{51} &    \ding{55}      &    \ding{55}         &         \ding{55}    & 35.076                \\
    \ding{51}                 & \ding{51} & \ding{51}      &    \ding{55}         &         \ding{55}    & 32.468                \\
    \ding{51}                 & \ding{51} & \ding{51}      & \ding{51}         &         \ding{55}    & 25.763               \\ 
    \ding{51}                 & \ding{51} & \ding{51}      & \ding{51}         & \ding{51}   & 25.043               \\ \bottomrule
    \end{tabular}
    \caption{Effect of each module of SphereDiffusion. `DRSE' / `DDaB' / `SR' / `SSCL' / `SGAG' represent our Distortion-Resilient Semantic Encoding / Deformable Distortion-aware Block / Spherical Reprojection / Spherical SimSiam Contrastive Learning / Spherical Geometry-aware Generation.}
    \label{tab:ablation}
\end{table}
\begin{figure*}[tb]
    \centering
    \includegraphics[width=0.98\linewidth]{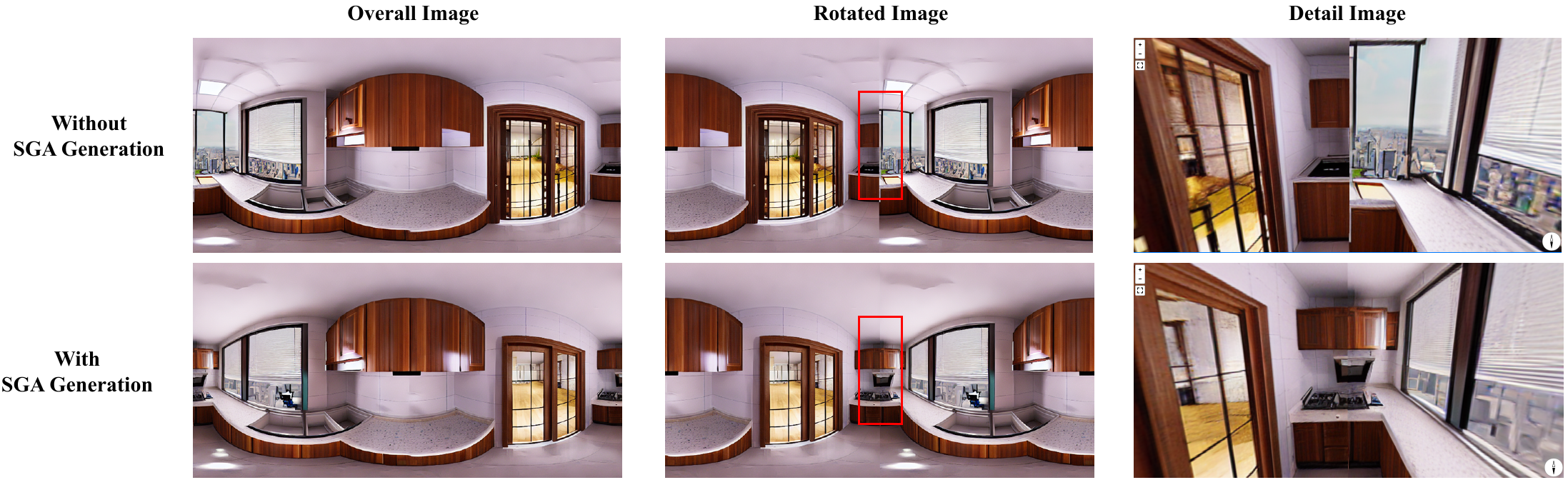}
    \caption{Visualization of generated image results with or without the Spherical Geometry-aware Generation. 
    We use the same SphereDiffusion model, employing consistent text prompts, segmentation maps, and random seeds for generation. 
    The first row shows images generated without incorporating SGA Generation, while the second row presents images generated with the inclusion of SGA Generation. `Rotated Image' is obtained by rotating the generated `Overview Image' by $\alpha = 180^{\circ}$.}
    \label{fig:vissgag}
    \centering
\end{figure*}
In this section, we compare the image generation quality with the latest work, and ~\Cref{tab:sota_comp} shows the performance comparison between our method and other approaches. 
To ensure a fair comparison, we use the official implementation of ControlNet with the same hyperparameters and training iterations on the Structured3D dataset. 
We select four FOV sizes, 30, 60, 90, and 120 degrees for comparison.
As can be seen, our method outperforms other methods in all metrics. 
Among them, in the test of four different FOV sizes, the most widely used FID score, our method improved significantly by 14.312 on average compared to ControlNet.

Furthermore, the visualization of the generated images also intuitively shows that our generation quality is more consistent with the textual descriptions and semantic segmentation maps compared to ControlNet. 
As shown in ~\Cref{fig:cmpsota}, when we want to generate a bedroom with white walls and a pink bed, we can see that ControlNet erroneously generates a room with pink walls, and the area originally labeled `curtain' in the semantic segmentation does not correctly generate the specified object. 
In contrast, since our method achieves a better object understanding of spherical panoramic images, our method accurately generates white walls and a pink bed, with the corresponding `curtain' area correctly generating the specified object, resulting in a more reasonable overall output. 
When we try to generate a kitchen with gray walls and white cabinets, ControlNet mistakenly generates gray cabinets. 
In contrast, our method correctly generates gray cabinets and gray walls.
Furthermore, the boundary connectivity of our generated images is significantly better than that generated by ControlNet.

\subsection{Ablation Study}
\subsection{Effect of Four Modules in Training Process}
As shown in ~\Cref{tab:ablation}, we validate Distortion-Resilient Semantic Encoding, Deformable Distortion-aware Block, Spherical SimSiam Contrastive Learning, and Spherical Reprojection, respectively.
We selected a FOV size of $90^{\circ}$ for the ablation experiment.
The baseline FID score is 39.450.
The FID score improves to 38.805, only including Distortion-Resilient Semantic Encoding, indicating the positive impact of incorporating semantic representation in the model.
Adding our Deformable Distortion-aware Block to the model with Distortion-Resilient Semantic Encoding further enhances the performance, achieving an improvement in the FID score of 3.7.
The combination of Distortion-Resilient Semantic Encoding, Deformable Distortion-aware Block, and Spherical Reprojection results in an improvement, with the FID score dropping to 32.468. 
This demonstrates the effectiveness of incorporating Spherical Reprojection in the model.
When we add all our components, the FID score further improves to 25.763, highlighting the importance of Spherical SimSiam Contrastive Learning in significantly enhancing the model's performance.
\subsection{Effect of Spherical Geometry-aware Generation}
We evaluate our SGA Generation through \Cref{tab:ablation} and \Cref{fig:sgag}.
As shown in ~\Cref{tab:ablation}, without retraining the model and only incorporating SGA Generation during the testing process, the FID score improves by almost 0.7. 
Visualizations are shown in ~\Cref{fig:sgag}. 
Without SGA Generation, the generated images exhibit discontinuity at the boundary, with a clear demarcation line. 
However, once using SGA Generation, the generated images exhibit better connectivity. 
This demonstrates that SGA Generation can use the spherical geometric characteristic to enhance the boundary connectivity of generated spherical panoramic images.
\section{Conclusion}
Generating spherical panoramic images is a challenging task, as it requires considering spherical distortion and geometric characteristics. 
We propose SphereDiffusion, a framework that accounts for these characteristics, generating high-quality, controllable spherical panoramic images from single NFOV segmentation maps and text prompts. 
For spherical distortion characteristic, we introduce Distortion-Resilient Semantic Encoding and Deformable Distortion-aware Block.
For spherical geometry characteristic, we leverage the spherical rotation invariance of spherical panoramic images and propose SGA Training, which includes Spherical Reprojection and Spherical SimSiam Contrastive Learning. 
Additionally, we introduce SGA Generation to improve the generation process.
Through experiments, we verified that our method can significantly improve the quality of the generated images.
\section*{Acknowledgements}
This work is supported in part by National Natural Science Foundation of China under Grant U20A20222, National Science Foundation for Distinguished Young Scholars under Grant 62225605, Zhejiang Key Research and Development Program under Grant 2023C03196, Research Fund of ARC Lab, Tencent PCG,  and sponsored by CCF-AFSG Research Fund as well as The Ng Teng Fong Charitable Foundation in the form of ZJU-SUTD IDEA Grant, 188170-11102.

\bibliography{aaai24}

\end{document}